\documentclass[10pt,twocolumn,letterpaper]{article}

\usepackage{iccv}
\usepackage{times}
\usepackage{epsfig}
\usepackage{graphicx}
\usepackage{amsmath}
\usepackage{amssymb}

\usepackage[breaklinks=true,bookmarks=false]{hyperref}

\iccvfinalcopy % *** Uncomment this line for the final submission

\def\iccvPaperID{****} % *** Enter the ICCV Paper ID here

\ificcvfinal\fi

\begin{document}

%%%%%%%%% TITLE
\title{Spatial and Temporal Networks for Facial Expression Recognition \\ in the Wild Videos}

\author{Shuyi Mao$^{1,*}$, Xinqi Fan$^{2,*}$, Xiaojiang Peng$^{1,*,\dag}$\\
$^{1}$Shenzhen Technology University, Shenzhen, China\\
$^{2}$City University of Hong Kong, Hong Kong SAR, China\\
{$^{*}$\small Equal contribution}\\
{$^{\dag}$\small Correspondence: \tt  pengxiaojiang@sztu.edu.cn}
}

% \and
% Xinqi Fan\\
% City University of Hong Kong\\
% \\
% {\tt\small xinqi.fan@my.cityu.edu.hk}
% }

\maketitle

\ificcvfinal\thispagestyle{empty}\fi

%%%%%%%%% ABSTRACT
%%%%%%%%%%%%%%%%%%%%%%%%%%%%%%%%%%%%%%%%%%%%%%%%%
\begin{abstract}
The paper describes our proposed methodology for the seven basic expression classification track of Affective Behavior Analysis in-the-wild (ABAW) Competition 2021. In this task, facial expression recognition (FER) methods aim to classify the correct expression category from a diverse background, but there are several challenges. First, to adapt the model to in-the-wild scenarios, we use the knowledge from pre-trained large-scale face recognition data. Second, we propose an ensemble model with a convolution neural network (CNN), a CNN-recurrent neural network (CNN-RNN), and a CNN-Transformer (CNN-Transformer), to incorporate both spatial and temporal information. Our ensemble model achieved $F_{1}$ as 0.4133, accuracy as $0.6216$ and final metric as $0.4821$ on the validation set. 
\end{abstract}

%%%%%%%%%%%%%%%%%%%%%%%%%%%%%%%%%%%%%%%%%%%%%%%%%
\section{Introduction}
Affect behavior analysis plays an important role in human-centered intelligence, where robots aim to understand human affects and respond to us smartly~\cite{kollias2021analysing}. There are many fields where affect behavior analysis can apply, such as health, education, marketing, etc~\cite{kollias2021analysing, kollias2020analysing}. However, analyzing affect behaviors in the wild is a challenging task, especially video-based ones. In Aff-Wild and Aff-Wild2~\cite{zafeiriou2017aff, kollias2019deep, kollias2019expression}, authors proposed the very first video-based affect behavior database, in which it contains 7 basic expressions, 2D valence-arousal, and 12 facial action units (AUs) annotations, to facilitate the society to solve these problems. Several multi-task approaches have been introduced to recognize these affective behaviors ~\cite{kollias2021distribution, kollias2021affect, kollias2019face}.

Among these tasks, facial expression recognition (FER) has more directly practical applications. The automatic FER systems can provide human-readable outputs as Anger, Disgust, Fear, Happiness, Sadness, Surprise, and Neutral. Many researchers proposed deep learning-based methods for FER in the wild~\cite{wang2020region, wang2020suppressing, li2018occlusion, li2018reliable}. However, rare studies focus on video-based FER in the wild. There are two major challenges in video-based FER in the wild. First, the environment is diverse and complex. Second, frames have temporal correlations.

In this work, we propose two methods to solve the above-mentioned problems. First, to adapt the model to in-the-wild scenarios, we use the knowledge from pre-trained large-scale face recognition data. Second, we propose an ensemble model with a convolution neural network (CNN), a CNN-recurrent neural network (CNN-RNN), and a CNN-Transformer, to incorporate both spatial and temporal information. Our ensemble model achieved $F_{1}$ as 0.4133, accuracy as $0.6216$ and final metric as $0.4821$ on the validation set. The source code of our work is publicly available online \footnote{https://github.com/xinqi-fan/ABAW2021}

The rest of this paper is organized as follows. We introduce the proposed method in Section 2, the experiments and discussion in Section 3, before concluding in Section 4.

%%%%%%%%%%%%%%%%%%%%%%%%%%%%%%%%%%%%%%%%%%%%%%%%%
\section{Methodology}
\begin{figure*}
\begin{center}
\includegraphics[width=0.85\textwidth]{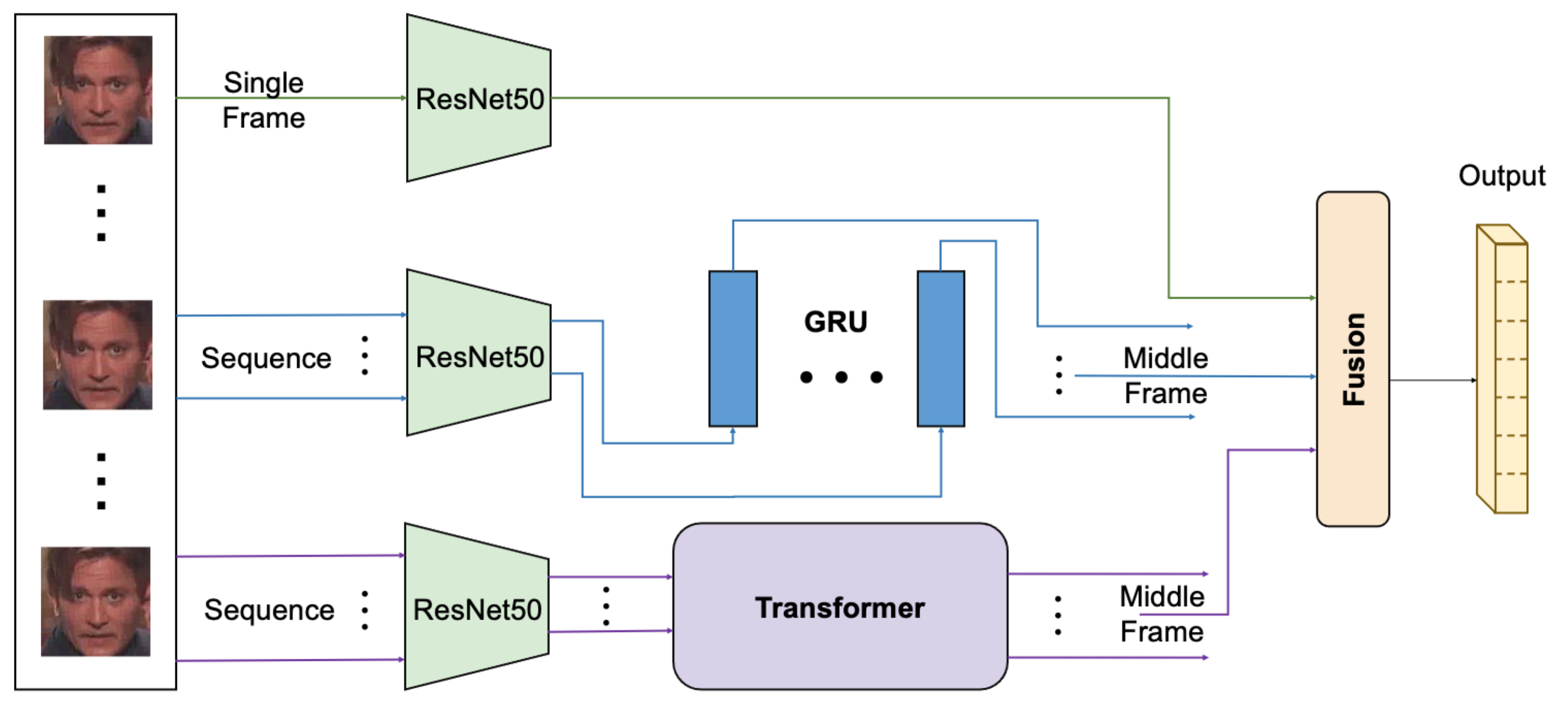}
\end{center}
   \caption{Pipeline of the proposed model.}
   %Each model was trained separately, and an ensemble model for them is used for final inference.
\label{fig:pipeline}
\end{figure*}

We propose an ensemble model (Fig.~\ref{fig:pipeline}), which contains a residual network with 50 layers (ResNet50) model, a CNN-RNN model, and a CNN-Transformer model, in which ResNet50 is used as the CNN. The ensemble model can incorporate both spatial and temporal information effectively. The ensemble is done through a weighted summation among each model.  

\subsection{ResNet50 with Pretrained on Large-scale Face Data}
ResNet50 was proposed in~\cite{he2016deep} to solve the training degradation problem. The network adds skip connections to allow gradients to smoothly passed back to early layers. To fully utilize the network's feature extraction ability, we use the pretrained weights from a large-scale face dataset VGGFace2~\cite{cao2018vggface2}, aiming to gain better diverse in-the-wild information and various faces. 

\subsection{Convolution Recurrent Neural Network}
To leverage the temporal information in the data, we first use CNNs to extract feature maps from consecutive frames, and then apply RNNs based on gated recurrent units (GRU)~\cite{cho2014learning}. In particular, our CNN is a ResNet50, and the depth of our GRU is 2.

\subsection{Convolution Transformer}
One disadvantage of RNNs is that it has to process frames in a sequential manner, which limits the information interaction and enlarges the training time. Transformers~\cite{vaswani2017attention} based on attention was proposed to solve such problems. The multi-head attention utilizes multiple queries, keys, and values to focus on the most related information at each query. We use the same sinusoidal positional encoding as the original transformers to combine the positional information into the network.

%%%%%%%%%%%%%%%%%%%%%%%%%%%%%%%%%%%%%%%%%%%%%%%%%
\section{Experiment and Result}
\subsection{Data}
Aff-Wild2 consists of 548 videos with 2, 813, 201 frames. 539 videos in Aff-Wild2 contain annotations in terms of the seven basic expressions. We use the provided cropped and aligned images, where a RetinaFace with ResNet face detector has been applied to crop the video into 112 $\times$ 112 faces and split into frames~\cite{kollias2021analysing}. 

\subsection{Implementation Detail}
We implement our model on PyTorch deep learning framework. We set the learning rate as 5e-4 and trained for 10 epochs with stochastic gradient descent (SGD) optimizer. A multi-step learning rate scheduler has been used at the 2, 4, and 8 steps. Our training platform is a high-performance computing center, where we use 4 GPU cards - NVIDIA 2080Ti with 11GB, 8 CPU cores. For temporal models, we train the networks in a many-to-many manner with 9 frames, but inference in a many-to-one way. The middle frame output is extracted out for the corresponding middle frame input for the ensemble.

\subsection{Evaluation Metric}
Evaluation metrics are $F_{1}$, total accuracy, and their combined final evaluation metric. $F_{1}$ is defined as 
\begin{equation}
    F_{1}=\frac{2 \times \text { precision } \times \text { recall }}{\text { precision }+\text { recall }}.
\end{equation}

The definiation of total accuray is
\begin{equation}
    \mathcal{T}Acc=\frac{\text { Number of Correct Predictions }}{\text { Total Number of Predictions }}.
\end{equation}

The final evaluation metric combines the above two with different weights for each as
\begin{equation}
    \mathcal{E}_{\text {total }}=0.67 \times F_{1}+0.33 \times \mathcal{T}Acc.
\end{equation}

\subsection{Result and Discussion}
The baseline model is a VGG-Face model with a customized fully connected layer for each task. We report our results on ResNet50, ResNet50 with VGGFace2 pretrained weights, CNN+CNN-RNN, CNN+CNN-Transformer, and the final CNN+CNN-RNN+CNN-Transformer (Table~\ref{tb:result_fer}). It shows that our full ensemble model achieves the best performance on all evaluation metrics. 

\begin{table}
\begin{center}
\begin{tabular}{|c|c|c|c|}
\hline 
\textbf{Model}                                                          & \textbf{$F_{1}$} & \textbf{$\mathcal{T}Acc$} & \textbf{$\mathcal{E}_{\text {total }}$} \\ \hline \hline
VGG-Face (Baseline)                                                     & 0.30             & 0.50                      & 0.36                                    \\ \hline
ResNet50                                                                & 0.3763           & 0.5713                    & 0.4402                                  \\ \hline
\begin{tabular}[c]{@{}c@{}}ResNet50 \\ (Pretrain VGGFace2)\end{tabular} & 0.4124           & 0.6216                    & 0.4815                                  \\ \hline
CNN+CNN-RNN                                                             & 0.4126           & 0.6220                    & 0.4817                                  \\ \hline
CNN+CNN-Transformer                                                     & 0.4127           & 0.6221                    & 0.4818                                  \\ \hline
\begin{tabular}[c]{@{}c@{}}CNN+CNN-RNN\\ +CNN-Transformer\end{tabular}  & \textbf{0.4133}           & \textbf{0.6216}                    & \textbf{0.4821}                                  \\ \hline
\end{tabular}
\end{center}
\caption{FER results on the evaluation dataset. Our ensemble model achieves the best final $\mathcal{E}_{\text {total }}$.}
\label{tb:result_fer}
\end{table}

In addition, we also validate our ResNet50 pretrained with VGGFace2 on the AU detection track. The validation is summarized in Table ~\ref{tb:result_au}. Details of evaluation metrics can be found in~\cite{kollias2021analysing}.

\begin{table}
\begin{center}
\begin{tabular}{|c|c|c|c|}
\hline 
\textbf{Model}                                                          & \textbf{$\mathcal{A} F_{1}$} & \textbf{$\mathcal{T}Acc$} & \textbf{$\mathcal{E}_{\text {total }}$} \\ \hline \hline
VGG-Face (Baseline)                                                     & 0.40             & 0.22                      & 0.31                                    \\ \hline
% ResNet50                                                                & 0.3763           & 0.5713                    & 0.4402                                  \\ \hline
\begin{tabular}[c]{@{}c@{}}ResNet50 \\ (Pretrain VGGFace2)\end{tabular} & 0.4701           & 0.8768                    & 0.6734                                                                    \\ \hline
\end{tabular}
\end{center}
\caption{AU results on the evaluation dataset.}
\label{tb:result_au}
\end{table}

%%%%%%%%%%%%%%%%%%%%%%%%%%%%%%%%%%%%%%%%%%%%%%%%%
\section{Conclusion}
In this work, we propose two methods to solve the challenging video-based FER in-the-wild problem. First, to adapt the model to in-the-wild scenarios, we use the knowledge from pre-trained large-scale face recognition data. Second, we propose an ensemble model with a ResNet50, a CNN-RNN, and a CNN-Transformer, to incorporate both spatial and temporal information. Our ensemble model achieved $F_{1}$ as 0.4133, accuracy as $0.6216$ and final metric as $0.4821$ on the validation set.

%%%%%%%%%%%%%%%%%%%%%%%%%%%%%%%%%%%%%%%%%%%%%%%%%
{\small
\bibliographystyle{ieee_fullname}
\bibliography{Reference_BibTex}
}
% Reference 
% \ifCLASSOPTIONcaptionsoff
%   \newpage
% \fi
% \section*{References and Footnotes}
% \bibliographystyle{IEEEtran}
% \bibliography{Reference_BibTex}

\end{document}